\documentclass{article}

    \PassOptionsToPackage{numbers, compress}{natbib}

\usepackage[preprint]{neurips_2024}

\usepackage[utf8]{inputenc} 
\usepackage[T1]{fontenc}    
\usepackage{url}            
\usepackage{booktabs}       
\usepackage{amsfonts}       
\usepackage{nicefrac}       
\usepackage{microtype}      
\usepackage{xcolor}         
\usepackage{graphicx}
\usepackage{subfigure}
\usepackage{amsmath}
\usepackage{amssymb}
\usepackage{mathtools}
\usepackage{amsthm}
\usepackage{multirow}

\usepackage{comment} 
\usepackage{longtable} 
\usepackage{tabularx} 
\usepackage{todonotes}
\usepackage{lipsum}
\usepackage{wrapfig}
\usepackage{placeins} 
\usepackage{pifont}

\makeatletter
\AtBeginDocument{%
  \renewcommand{\sectionautorefname}{\S\@gobble}

  \renewcommand{\subsectionautorefname}{\S\@gobble}  
}
\makeatother

\newcommand\rurl[1]{%
  \href{https://#1}{\nolinkurl{#1}}%
}
\definecolor{mydarkblue}{rgb}{0,0.08,0.45}
\usepackage[colorlinks,citecolor=mydarkblue,urlcolor=mydarkblue,linkcolor=mydarkblue]{hyperref}
\usepackage{cleveref}

\usepackage{minitoc} 

\noptcrule

\title{Critical Data Size of Language Models \\ from a Grokking  Perspective}

\author{
\vspace{2mm} 
        Xuekai Zhu~$^1$\hspace{9mm}
        Yao Fu~$^3$ \hspace{9mm}
        Bowen Zhou~$^2$ \hspace{9mm}
        Zhouhan Lin~$^1$
    \\
    \hspace{-3mm}
    \textsuperscript{1} Shanghai Jiao Tong University  
    \hspace{4mm}  
    \textsuperscript{2} Tsinghua University 
    \hspace{4mm}  
    \textsuperscript{3} University of Edinburgh
    \vspace{2mm}
    \\
    \tt \href{mailto:xuekaizhu0@gmail.com}{xuekaizhu0@gmail.com} 
    \hspace{4mm}
    \href{https://github.com/Xuekai-Zhu/Critical-Data-Size}{https://github.com/Xuekai-Zhu/Critical-Data-Size} 
}
    \vspace{4mm}

\begin{document}

\maketitle

\begin{abstract}
  We explore the critical data size in language models, a threshold that marks a fundamental shift from quick memorization to slow generalization. We formalize the phase transition under the grokking configuration into the Data Efficiency Hypothesis and identify data insufficiency, sufficiency, and surplus regimes in language models training dynamics. We develop a grokking configuration to reproduce grokking on simplistic language models stably by rescaling initialization and weight decay. We show that generalization occurs only when language models reach a critical size. We analyze grokking across sample-wise and model-wise, verifying the proposed data efficiency hypothesis. Our experiments reveal smoother phase transitions occurring at the critical dataset size for language datasets. As the model size increases, this critical point also becomes larger, indicating that larger models require more data. Our results deepen the understanding of language model training, offering a novel perspective on the role of data in the learning mechanism of language models. 
\end{abstract}

\begin{figure*}[!t]
\small
  \centering
  \includegraphics[width=\linewidth]{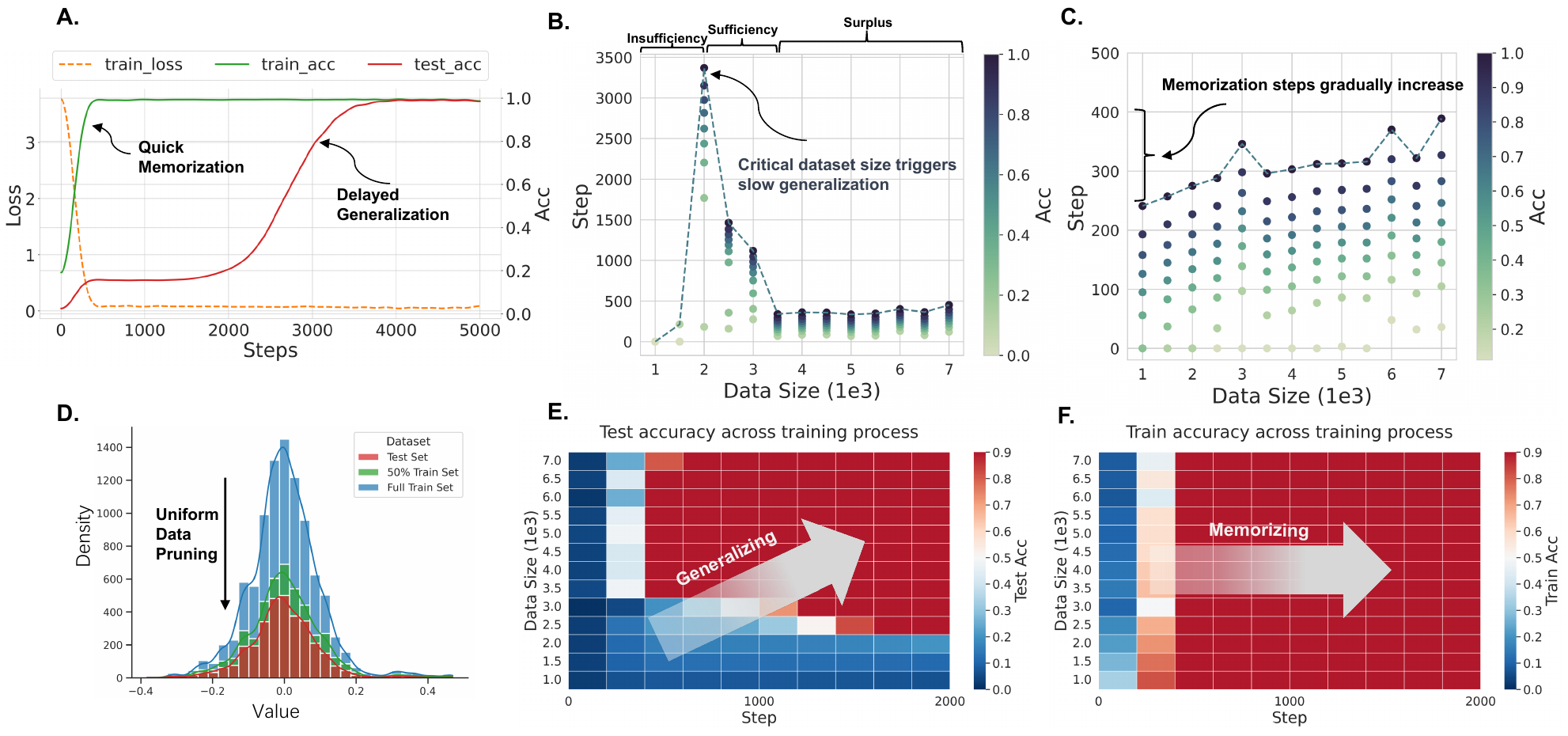}
  \caption{Comprehensive analysis of training dynamics and accuracy curves verifies the data efficiency hypothesis on vanilla grokking~\cite{power2022grokking,varma2023explaining}.
  \textbf{A:} Reproduced grokking phenomenon on modular addition 
   using a 1-layer decoder-only Transformer trained on 2000 samples. Delayed generalization ($\approx100\%$ test acc) occurs during continuous training after memorization completion ($\approx100\%$ train acc, overfitting).
  \textbf{B:} Step-wise Analysis of Test Accuracy. We observe a clear peak indicating slow generalization at the critical data size, while more training samples markedly speed up generalization. Below the critical data size, no generalization happens.
  \textbf{C:} Step-wise Analysis of Training Accuracy. Within 400 steps, the model can memorize all training data. Across various dataset sizes, there is a very small difference in memorization steps. 
  \textbf{D:} 1D PCA visualization of modular addition datasets. Data pruning uniformly samples from the initial distribution.
  \textbf{E} and \textbf{F:} Test / Training accuracy across the whole training process. The detailed training process is presented in Figure~\ref{fig:diff_data_size_on_modular}.
  }
  \label{fig:cds_on_modular}
\end{figure*}

\section{Introduction}
\label{sec:intro}
How does large data drive language models from memorization to generalization? 
Aiming to understand the generalization mechanism, researchers have made a series of striking discoveries across generalization abilities, including neural scaling laws~\cite{kaplan2020scaling}, double descent~\cite{nakkiran2021deep}, grokking~\cite{power2022grokking}
and emergent abilities~\cite{wei2022emergent}. Combining with recent powerful large language models~(LLMs)~\cite{touvron2023llama, Achiam2023GPT4TR, Anil2023GeminiAF}, we get a solid wisdom: \textit{Generalization comes from large data.} The complex interplay between data and generalization compels us back to the heart of machine learning, a fundamental puzzle about the role of data. 

This paper studies the question of how data scaling influences generalization by investigating the grokking phenomenon, and our answer is that \textit{there is a critical data size inside the learning process of language models, driving language models from memorizing to generalizing}.
Specifically, once we pass critical data size, the model transitions toward generalization. As illustrated in Figure~\ref{fig:cds_on_modular}, with insufficient data, the language model only memorizes training data. 
As the data scale approaches a critical size, slow generalization happens.
After the data size progressively exceeds this critical size, the model not only memorizes but also generalizes more effectively, demonstrating a swift convergence in its learning process. We formulate these findings into the \textit{Data Efficiency Hypothesis}. Furthermore, the sample-wise grokking conforms to this hypothesis on modular addition, IMDB, and Yelp datasets. Model-wise grokking provides evidence that the critical data size increases with model sizes. The case study on instruction tuning validates our hypothesis on more complex tasks. Note that all results are observed under a simplistic grokking configuration expanding from ~\cite{liu2022omnigrok}\footnote{\citet{liu2022omnigrok} proposed that rescaling initialization parameters can induce grokking on MNIST and IMDB using MLPs and LSTM, respectively.}.

Recent discoveries have partially explained the role of the data through high-quality data pruning and grokking-related data efficiency analysis.
\citet{sorscher2022beyond} reproduced exponential scaling law via data pruning, indicating many highly redundant examples in training sets. \citet{power2022grokking} firstly identified the grokking phenomenon, delayed generalization after overfitting, and coupled generalization with the data size~\cite{liu2022towards,liu2022omnigrok}. 
\citet{varma2023explaining} uncovered a crossover point where generalization becomes more efficient than memorization, revealing semi-grokking and ungrokking. 
Building upon these insights, we investigate the critical data size that triggers generalization. Our approach involves an expansion of data-dependent grokking and a detailed analysis of the learning phase transition in language models.


Our work deepens the understanding of language models' data efficiency and grokking from three perspectives: 

\noindent\textbf{I.} We introduce the critical data size in language models under the grokking configuration, a distinct threshold marking the phase transition between memorization and generalization. We formalize the data efficiency hypothesis to define the critical data size in language models precisely. (\S~\ref{sec:grokking_law});

\noindent\textbf{II.} By rescaling parameters initialization and weight decay, we construct a method to stably build grokking phenomena on \textit{real} language model tasks (Yelp and IMDB, \S~\ref{sec:experimental_setup}), a step forward from existing work which is usually on \textit{synthetic} settings, e.g., modular addition;
    
\noindent\textbf{III.} We thoroughly verify the data efficiency hypothesis across sample-wise grokking~(\S~\ref{sec:sample_wise_grokking}) and model-wise grokking~(\S~\ref{sec:model-wise_grokking}). Additionally, we provide a detailed phase analysis in language models during the grokking process~(\S~\ref{sec:learning_process_analysis}) and an expanded case study on the instruction tuning task~(\S~\ref{sec:case_study}).

\section{Data Efficiency Hypothesis}
\label{sec:grokking_law}
Following the theory framework in~\cite{nakkiran2021deep}, we formulate a hypothesis about critical data size under grokking, which demonstrates that generalization depends on effective data size.

We introduce the concept of \textbf{\textit{Critical Data Size}} (\textbf{CDS}) for grokking-related model training. Given a model $ \mathcal{M}$ and a data distribution $\mathcal{D}$, the \textit{Critical Data Size} denotes the minimum number of samples $n$ required to achieve approximately 100\% training accuracy on $S_{train}$ and a converged test accuracy $\epsilon$ on $S_{test}$.

\textbf{Definition 1 (Critical Data Size)}
\textit{For a given data distribution $ \mathcal{D} $, model $ \mathcal{M} $, and converged test performance $\epsilon$, the CDS is}:
\begin{equation}
\operatorname{CDS}(\mathcal{D}, \mathcal{M}) \coloneqq \min \left\{n \mid \mathbb{E}_{S \sim \mathcal{D}^n}\left[\operatorname{Acc}_S(\mathcal{M})\right] \geq \epsilon\right\},
\end{equation}
\textit{where $Acc_S(\mathcal{M})$ is the mean accuracy of model $\mathcal{M}$ on test samples $S_{test} \in S$.}

\textbf{Hypothesis 1 (Data Efficiency Hypothesis, informal)}
Given any natural data distribution $\mathcal{D}$ and a model $\mathcal{M}$ with a convergence threshold $ \epsilon $, when predicting labels based on $n$ samples drawn from $\mathcal{D}$, the following regimes are proposed:

\hspace{1em}\textbf{Data Insufficiency Regime}
When $ n $ is significantly less than $\operatorname{CDS}(\mathcal{D}, \mathcal{M})$, the performance of model $ \mathcal{M} $ might not reach optimality, irrespective of the complexity of the model. Suggesting: memorization only, no generalization.

\hspace{1em}\textbf{Data Sufficiency Regime}
When $ n\approx \operatorname{CDS}(\mathcal{D}, \mathcal{M})$, model $ \mathcal{M} $ will approach the desired optimal performance level $ \epsilon $. We will observe delayed generalization, and further data addition may only accelerate the convergence without enhancing performance. Suggesting: memorization, then delayed generalization -- grokking.

\hspace{1em}\textbf{Data Surplus Regime}
When $n$ significantly surpasses $\operatorname{CDS}(\mathcal{D}, \mathcal{M})$, 
extra data will accelerate the convergence, manifesting as simultaneous memorization and generalization.
Suggesting: concurrent memorization and generalization -- no grokking.

Our hypothesis is informal,  as it depends on prior knowledge of the model's optimal performance $\epsilon$ and the current dataset distribution $\mathcal{D}$. We have yet to formally define the concepts of `Insufficiency,' `Sufficiency,' and `Surplus,' which are contingent upon the practical context of the initial dataset and the experimental model. However, under a specific and determined setting, we are able to observe these phase changes through experimental phenomena.

\section{Grokking Configuration for Language Models}
\label{sec:experimental_setup}
\subsection{Configuration to Induce Grokking }
As aforementioned, the grokking is heavily coupled with data sizes. We try to investigate the role of data by inducing the grokking in language models. 
To this end, we introduce a grokking configuration derived from Omnigrok~\cite{liu2022omnigrok}. In the grokking configuration, each weight of models is rescaled by a factor $\alpha$, derived as follows:
\begin{align}
\mathbf{w} = \alpha * \mathbf{w_0},~~\text{where}~\alpha \equiv \frac{\|\mathbf{w}\|_2 }{\|\mathbf{w_0}\|_2},
\label{eq:rescale_initialization}
\end{align}
and $\mathbf{w_0}$ represents standard Pytorch initialization of model parameters. 
The parameter $\alpha$ serves the purpose of amplifying the grokking phenomenon, as is suggested by~\citet{liu2022omnigrok}.
Then, we optimize rescaled models $\mathbf{w}$ by cross-entropy loss and weight decay. Given a set of inputs $\mathbf{X}$, a set of labels $\mathbf{Y}$ and a training dataset $\mathcal{D}=\{(x_1,y_1), (x_2,y_2),..., (x_n,y_n)\}$, the optimization objective is formalized as follows:

\begin{align}
&\mathcal{L}(\mathbf{w}) = \mathcal{L}_{\text{CE}}(\mathbf{w}) + \mathcal{L}_{\text{WD}}(\mathbf{w}), \label{learning_objective}
\\
&\mathcal{L}_{\text{CE}}(\mathbf{w}) = -\frac{1}{n} \sum_{(x,y) \in \mathcal{D}} \log \left( \frac{\exp(h(x, y))}{\sum_{y' \in \mathbf{Y}} \exp(h(x, y'))} \right),
\\
&\mathcal{L}_{\text{WD}}(\mathbf{w}) = \frac{\lambda}{2} \|\mathbf{w}\|_2,
\end{align}

where \( \mathcal{L}_{\text{CE}}(\mathbf{w}) \) represents the cross-entropy loss, and \( \mathcal{L}_{\text{WD}}(\mathbf{w}) \) denotes the weight decay. 

Previous studies have demonstrated that generalization significantly depends on regularization~\cite{power2022grokking,doshi2023grok}. In our approach, we set the dropout ratio to 0, indicating that the learning process of the model 
$\mathbf{w}$ is exclusively influenced by cross-entropy loss and weight decay. As illustrated in Figure~\ref{fig:cds_on_modular}A, the training loss ﬁrst quickly drops to 0, achieving training set memorization. Since the gradient of the training loss is also near zero, without any regularization, the model weights would remain static (i.e., overfitting). Consequently, when the training loss approaches zero, weight decay becomes the only source that drives further model optimization, leading to generalization. Formally, we formalize the weight decay as follows:
\begin{align}
w(t) \approx \exp(-\gamma t)w_0,
\end{align}
where $w=\|\mathbf{w}\|_2$, $w_0=\|\mathbf{w}_0\|_2$, and $\gamma$ is the decay constant, which determines the decay rate. The generalization time $t$ can be derived as: 
\begin{align}
t \approx \frac{\ln(w_0/w^*)}{\gamma},
\end{align}
where $ t\propto \gamma^{-1}$, and $w^*$ is the L2 norm of optimal weights in generalization. 

In summary, as discussed in~\cite{power2022grokking,liu2022omnigrok}, huge initialization $\alpha$ and small weight decay $\gamma$ leads to a delay generalization, i.e., grokking. We use these two factors to induce grokking on language datasets. The detailed implementation setting is presented in Appendix~\ref{sec:training_details}.


\subsection{Training Dynamics under Data Pruning}
We verify our data efficiency hypothesis by analyzing training dynamics and seeing insights through data pruning. Specifically, we employ uniform data pruning to investigate the data dependence in grokking, which is a simple but effective analysis method to disentangle data dependence. The uniform data pruning randomly selects \( n \) elements from a dataset of size \( N \) without replacement, each with an equal probability of \( \frac{1}{N} \). This approach ensures a statistically uniform subset, maintaining the original dataset's proportional characteristics.
As illustrated in Figure~\ref{fig:cds_on_modular}D and~\ref{fig:data_pruning_imdb_pca}, uniform data pruning obtains a sub-dataset by proportionally extracting from the source distribution (i.e., distribution of the training set). More details are provided in Appendix~\ref{sec:Training_Dynamics}.



\subsection{Datasets and Implementation}\label{sec:Datasets_and_Implementation}
Here, we provide an overview of our experimental setup. Statistics of the datasets and a detailed comparison of experimental settings are described in Table~\ref{tab:data_statistics} and Table~\ref{tab:grokking_datasets}. More training details are presented in Appendix~\ref{sec:training_details}. 

\paragraph{Datasets}
We include four datasets in our experiments: modular addition~\citep{power2022grokking}, IMDB~\citep{maas-EtAl:2011:ACL-HLT2011}, Yelp~\citep{zhangCharacterlevelConvolutionalNetworks2015} and Natural-Instructions~\cite{wang-etal-2022-super}. 
Following~\citet{varma2023explaining}, we construct the modular addition using the formula a+b\%p=n. Specifically, we set p to 113. This dataset serves as a basic model for vanilla grokking studies. In contrast, IMDB and Yelp are well-known text classification datasets, offering more realistic samples than the synthetic modular addition dataset. The instruction tuning task is used as validation for more complex setting, which closely resembles real-world human usage scenarios. 

\paragraph{Implementation}
For vanilla grokking experiments, we employ a 1-layer decoder-only Transformer, similar to the models used in \cite{power2022grokking} and \cite{varma2023explaining}.
We set $\alpha$ and weight decay to $0$ and $1$, respectively.
For grokking on the IMDB and Yelp datasets, we opt for a 1-layer encoder-only Transformer, which is a more common setting for text classification tasks. We set $\alpha$ and weight decay $\gamma$ to 10 and 1e-2, respectively. All experiments are conducted on NVIDIA GeForce RTX 3090 GPU by training from scratch. 
Each reported result can be reproduced within three hours using a single GPU.

\section{Sample-wise Grokking}
\label{sec:sample_wise_grokking}

In this section, we report the main results of the simplistic Transformer-style models on modular addition, IMDB, and Yelp under the grokking configuration. 
We analyze the training dynamics and test accuracy, focusing particularly on how varying data sizes influence the learning process across diverse tasks when training models to completion.

\subsection{Vanilla Grokking}
To validate our proposed data efficiency hypothesis in vanilla grokking, we visualized training and test set accuracies with increasing training samples $N$ in modular addition.
As illustrated in Figure~\ref{fig:cds_on_modular}, 
we can see the grokking (i.e., slow generalization) is heavily data-size dependent. These results reveal: \textbf{(1) Generalization occurs at the critical data set size.} As shown in Figure~\ref{fig:cds_on_modular}B, there is a distinct peak, where the model slowly arrives at $100\%$ test accuracy. Specifically, upon reaching approximately 3500 steps, the model achieves perfect generalization, reflected in 100\% test accuracy.  Interestingly, the model has already memorized all training data by this point, reaching 100\% training accuracy within the first 500 steps. When the dataset reaches this critical size, the grokking phenomenon becomes notably apparent, as illustrated in Figure~\ref{fig:cds_on_modular}A.
\textbf{(2) Memorization and generalization occur almost simultaneously when the dataset is sufficient.} As data sizes pass the critical data size, generalization steps significantly decrease from 3500 to 500. This aligns closely with the number of steps necessary to fully learn all training samples.
\textbf{(3) Generalization cannot occur with insufficient data.} Below the critical data size, the model can only memorize the training samples but cannot generalize to test samples. For instance, as shown in Figure~\ref{fig:cds_on_modular}E, 1000 and 1500 training samples cannot even achieve 50\% accuracy (blue rows). Conversely, in Figure~\ref{fig:cds_on_modular}F, the number of memorization steps is almost unaffected by the dataset size.

\subsection{Grokking on IMDB}
\begin{figure*}[ht] 
\centering 
\includegraphics[width=\linewidth]{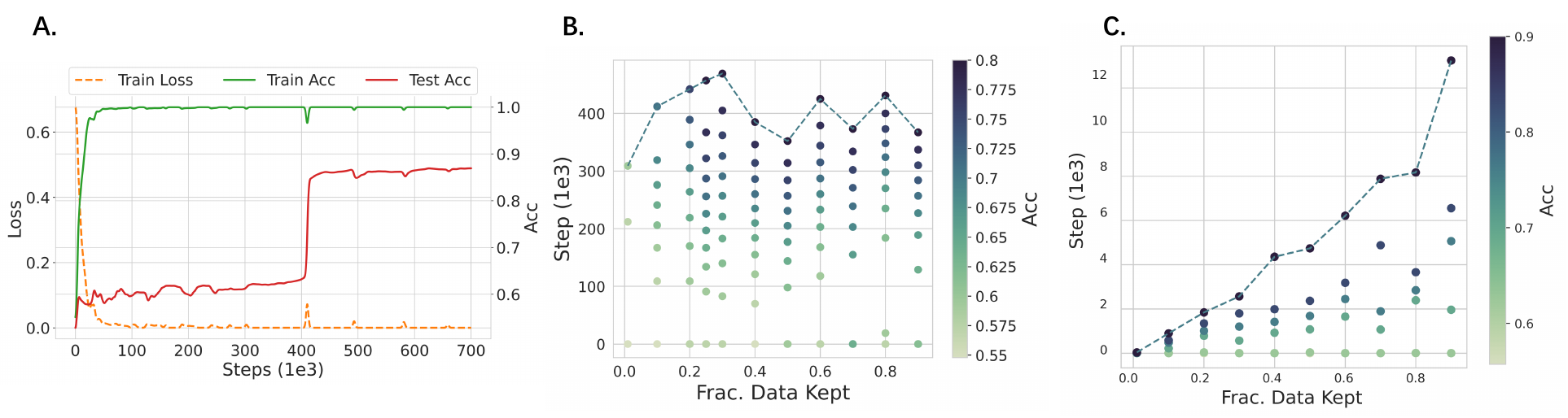} 
\caption{
\textbf{A:} We induce the grokking phenomenon on IMDB~\cite{maas-EtAl:2011:ACL-HLT2011} using a 1-layer encoder-only Transformer. The model suddenly flipped from memorizing training data to generalizing unseen test data after training for much longer. 
\textbf{B:} Step-wise Analysis of Test Accuracy in IMDB. We can observe a growth of steps, moving from very weak generalization (the light dot) to full-fledged generalization (the dark dot). As data fraction increases, generalization steps rapidly reach and maintain stability. 
\textbf{C:} Step-wise Analysis of Training Accuracy in IMDB. The memorization steps gradually increase with the fraction of the dataset. And it can finally achieve 100\% accuracy on the training data.
}
\label{fig:grokking_on_imdb} 
\end{figure*}
As previously discussed, through the grokking configuration, we successfully build the grokking phenomenon on the IMDB dataset~\cite{maas-EtAl:2011:ACL-HLT2011} using a single-layer, encoder-only Transformer. This is achieved by strategically rescaling parameter initialization and weight decay. As illustrated in Figure~\ref{fig:grokking_on_imdb}A, the training loss (yellow line) rapidly diminishes to zero, while the training accuracy nears 100\%. On the contrary, the test accuracy remains at the chance level ($\approx$50\%). Upon continual training for about 350,000 steps, a sudden leap in generalization occurs, and the model's accuracy jumps to about 87\% accuracy. These observations are consistent with the initial grokking~\cite{power2022grokking} on the modular addition task. However, given that the IMDB dataset is more complex than modular addition, the model cannot generalize perfectly.

Under the grokking configuration, we visualize the phased learning process across various fractions of the whole dataset. These
results reveal: \textbf{(1) the data efficiency hypothesis also applies to language models learning from the IMDB dataset.} As depicted in Figure~\ref{fig:grokking_on_imdb}B, there is a clear peak of generalization steps, indicating the critical data size. Before this threshold, the model achieves weak generalization ($\approx$67\%); as the data fraction grows, it gradually achieves high-quality generalization ($\approx$80\%). As the critical data size surpasses, the generalization steps stay around 400,000. As shown in Figure~\ref{fig:grokking_on_imdb}C, the memorization steps in training increase with the data size increases. This is obvious, as the larger the dataset, the more samples to be memorized.
\textbf{(2) The memorization and generalization phase change is smoother in realistic datasets.} Compared with vanilla grokking, phase transition on the IMDB is less abrupt and more gradual in nature.

\paragraph{Discussion: why is the phase transition in language datasets smoother?} 
We speculate there are two main reasons: the initial data size and task complexity. Larger data sizes can lead to de-grokking~\cite{liu2022omnigrok,varma2023explaining}. The model, training from a larger initial dataset, can learn a greater number of correlations. Consequently, the relationship between the phase transition and dataset size becomes increasingly smooth. The ``Effective Theory of Representation Learning''~\cite{liu2022towards} suggests that tasks such as modular addition require more refined (linear) representations. However, representations for language tasks tend to be significantly more complex or ``messier'' (i.e., nonlinear).

\begin{figure*}[ht] 
\centering 
\includegraphics[width=\linewidth]{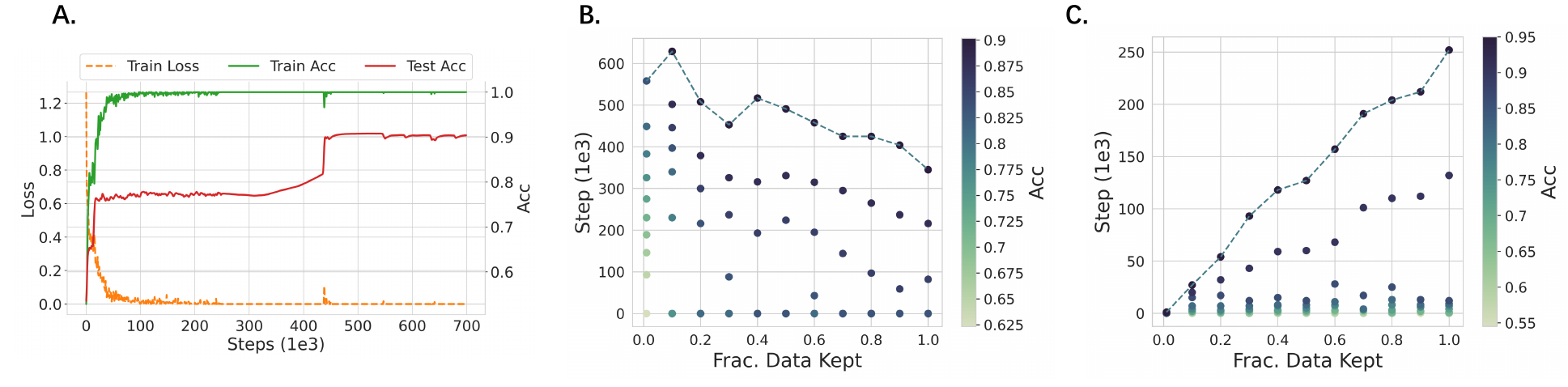} 
\caption{
\textbf{A:} We employ a 1-layer, encoder-only  Transformer to trigger the grokking phenomenon on $10\%$ Yelp data~\cite{zhangCharacterlevelConvolutionalNetworks2015}. The delayed generalization occurs after overfitting.
\textbf{B:} Step-wise Analysis of Test Accuracy in Yelp. The generalization steps first increase and subsequently decrease as the data fraction grows, which is consistent with results on modular addition and IMDB datasets.
\textbf{C:} Step-wise Analysis
of Training Accuracy in Yelp. Similar to experiments of modular addition and IMDB datasets, we obtain the same conclusion: memorization steps increase as the dataset size expands.
The detailed training process is presented in Figure~\ref{fig:acc_on_yelp}.
}
\label{fig:grokking_on_yelp} 
\end{figure*}
\subsection{Grokking on Yelp}
As shown in Figure~\ref{fig:grokking_on_yelp}A, we successfully induced the grokking phenomenon on Yelp. This was achieved using a smaller, 10\% subset of the Yelp dataset. 
As illustrated in Figure~\ref{fig:acc_on_yelp},
we discovered that using larger datasets could lead to `de-grokking', i.e., memorization and generalization co-occur. To promote grokking in the Yelp dataset, we strategically pruned the data under the grokking configuration.
Under various fractions of Yelp training samples, the results align with the proposed data efficiency hypothesis. This is evident in Figures~\ref{fig:grokking_on_yelp}B and ~\ref{fig:grokking_on_yelp}C, where we observe two fundamental phenomena: \textbf{(1) The presence of critical data size.} However, the transition from slow to faster generalization has become smoother.  As illustrated in Figure~\ref{fig:acc_on_yelp},  reducing the data size from 100\% to 10\% results in only about a 5\% decrease in performance. \textbf{(2) A trend where increasing the training sample size leads to a decrease in generalization steps.} Compared with IMDB and modular addition datasets, the Yelp dataset contains more samples, which leads to faster model fitting. Additional data did not lead to an improvement, but it did accelerate convergence. On the contrary, faster fitting weakens the manifestations associated with the data efficiency hypothesis, thus the phenomena we observe on Yelp are somewhat less pronounced.

\begin{figure*}[ht] 
\centering 
\includegraphics[width=\linewidth]{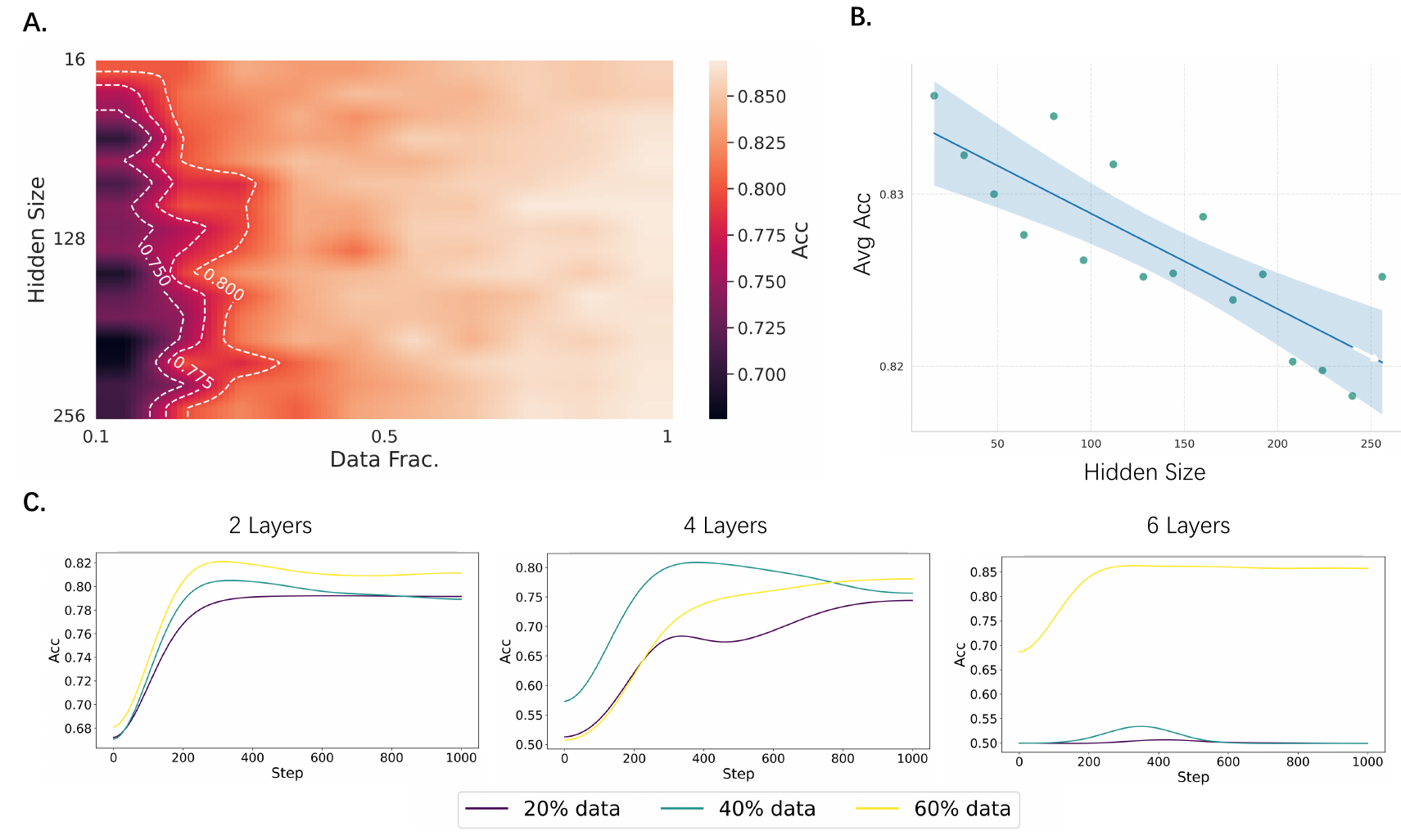} 
\caption{
Model-wise grokking experiments on IMDB demonstrate that the critical data size increases as the model size increases.
\textbf{A:} Test accuracy variations by hidden layer size and data fraction of the IMDB dataset. The data fraction required for higher accuracy increases as the model size increases. Training acc visualization is presented in Figure~\ref{fig:model_wise_grokking_train_acc}.
\textbf{B:} Average accuracy across all data fractions from 10\% to 100\%. The white arrows indicate that the average accuracy decreases as the model size increases, suggesting that larger models require more data to maintain performance. The light blue area represents a 95\% confidence interval.
\textbf{C:} 
Training curves for models with different layer counts under various data fractions. As the number of layers increases, larger models require larger data sizes for effective generalization.
}
\label{fig:model_wise_grokking} 
\end{figure*}

\begin{figure*}
\centering 
\includegraphics[width=0.98\linewidth]{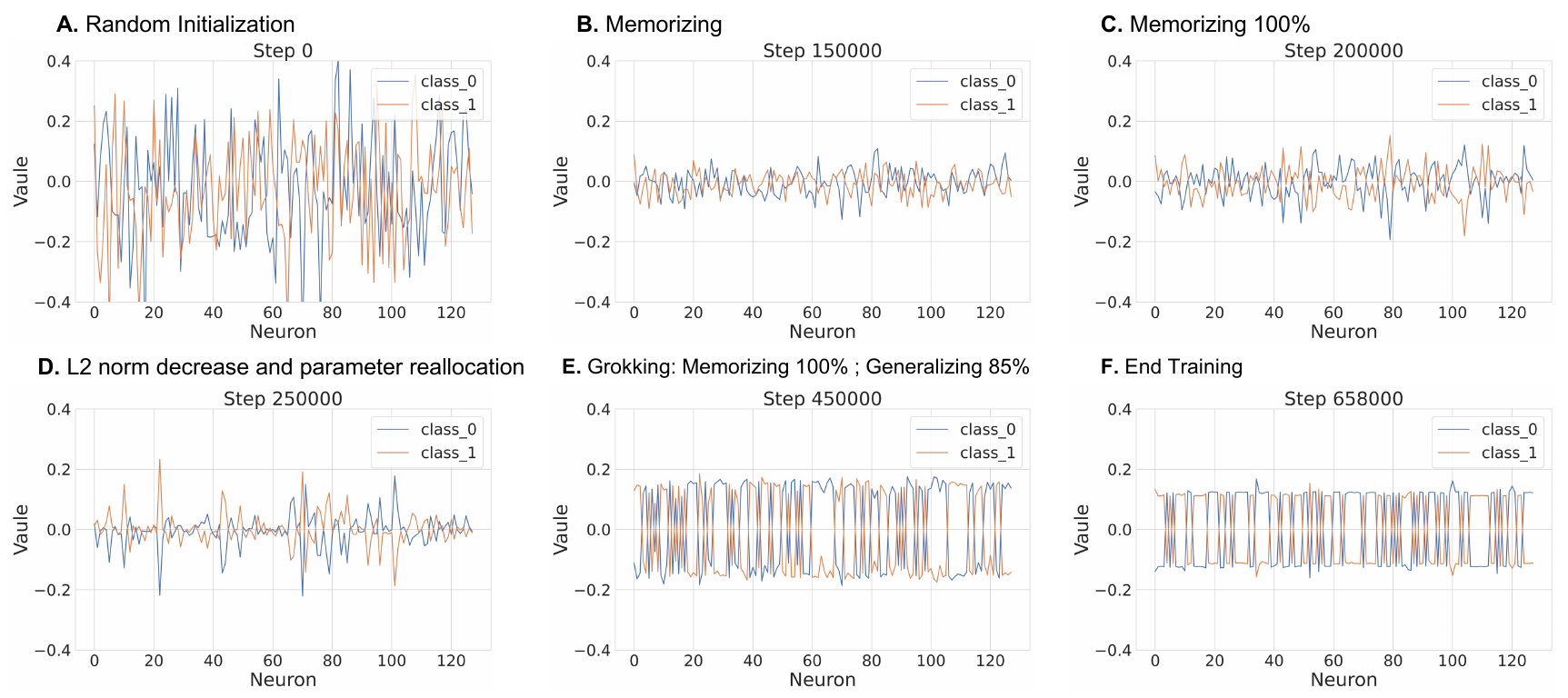} 
\caption{
Visualization about how the model transits from memorization to generalization throughout the training process.
We visualize the classification layer's weights during the learning process using a 1-layer, encoder-only Transformer on the IMDB dataset. 
Notably, the parameter distribution evolves from a randomly initialized state to a fixed range of values, which we have categorized into stages from A to F.
The transition from memorization to generalization is influenced by weight decay and loss, leading to a decrease in the L2 norm. More explanations of the L2 Norm evolution are in Figure~\ref{fig:grokking_stages}.
}
\label{fig:imdb_weights_analysis} 
\end{figure*}

\paragraph{Discussion: why grokking is not commonly observed in large language models with big datasets?
} 
(1) The model converges faster with larger datasets. If we have a `magnifying glass' to observe the learning process carefully, perhaps we can witness the grokking phenomenon in LLMs. Specifically, our experiments suggest that grokking can be more readily induced through strategic data pruning and grokking configuration. This approach essentially represents a `slow' learning version~(i.e., reduce dataset size, increase initialization, decrease weight decay) of modern learning systems. 
From this perspective, we conjecture that grokking is a fundamental phenomenon hidden under complex conditions, which can only be seen under the dual effects of dataset pruning and grokking configuration. 
(2) Large language models incorporate a variety of regularization methods, while our grokking simplistic model is limited to using only weight decay. As is well known, various regularization techniques in modern large models help accelerate convergence and prevent overfitting. In our simplified setting, the model's convergence speed is slowed down, allowing us to observe clear phase changes.

\section{Model-wise Grokking}
\label{sec:model-wise_grokking}

\paragraph{Experiments Setup} In this section, we present model-wise grokking experiments conducted on the IMDB dataset.
We aim to investigate the correlation between critical data size and various model sizes.
To achieve this, we systematically vary the models' hidden sizes and number of layers for a comprehensive analysis.
We adjust the model's hidden size from 16 to 256 and train on proportionally increasing subsets. The interval between each model size is 16. Each subset of the original dataset ranges from 10\% to 100\%, in increments of 10\%. Similarly, we vary the number of layers from 2 to 6 across different data sizes.
The 1D PCA visualization of data fractions of IMDB is presented in Figure~\ref{fig:data_pruning_imdb_pca}.
We visualize the accuracy landscapes across hidden sizes and data fractions, as well as the training curves for models with different numbers of layers. The main results reported in Figure~\ref{fig:model_wise_grokking} and~\ref{fig:model_wise_grokking_train_acc}.

\paragraph{Results: The critical data size increases with model size.}
As shown in Figure~\ref{fig:model_wise_grokking}A, a clear trend is observed: as the hidden layer size increases from 16 to 256, the area of low accuracy ($\leq 0.8$) gradually increases. 
The delineated contour line represents the performance around the critical data size.
This suggests a threshold beyond which increasing the model complexity does not yield proportional gains in performance. 
This pattern implies that larger models may require more data to achieve similar levels of accuracy, compared to smaller models. 
On the other hand, the delineated contour line also indicates the optimal combinations of hidden sizes and data fractions.
It's clear that the highest accuracies are concentrated in the lower-right quadrant, corresponding to larger data fractions and model sizes.
Figure~\ref{fig:model_wise_grokking}B
illustrates the relationship between the average accuracy (Avg Acc) and the hidden layer size. The average accuracy is the mean performance across all data fractions.
The results reveal a downward trend, indicating that the average accuracy decreases as the hidden size increases. 
In general, under fixed training data sizes, larger hidden sizes may not improve performance and can even damage the performance, while optimal accuracy is achieved at more minor scales. 
Figure~\ref{fig:model_wise_grokking}C further supports these findings: as the number of layers increases, larger models need more data to generalize effectively.

\paragraph{Discussion: Under the grokking configuration, larger models may not always beat smaller ones.}
According to neural scaling law~\cite{kaplan2020scaling,hoffmann2022training}, reaching the same level loss, the smaller model needs more data. In other words, large models can achieve better performance under the same data sizes. But things are different in our simplistic setting. 
As shown in Figure~\ref{fig:model_wise_grokking}A, some smaller models achieve better performance than slightly larger models when trained with smaller datasets. These observations are similar to inverse scaling~\cite{mckenzie2023inverse}, indicating worse performance with increased scale. 
From this perspective, we speculate that: 
(1) Learning real language tasks under the simplistic grokking configuration is challenging, similar to an inverse-scaling task. Since only using 1-layer parameters and weak regularization, the grokking configuration has limited capabilities on real-world language datasets. However, training LLMs often involves multi-task learning and huge parameters, which can significantly enhance the individually challenging task. While conventional wisdom suggests that increased model size generally improves performance, inverse scaling indicates a more complex interaction between model size and data sufficiency, i.e., the critical data size increases with model size. On the contrary, larger models do not always guarantee superior results when the available data does not meet the critical data size. 
This finding emphasizes balancing model size with data quantity in learning a single challenging task.
(2) Effective generalization in learning single challenging tasks may require a minimum necessary training set size, termed as the critical data size. As shown in Figure~\ref{fig:model_wise_grokking}A, the lighter area on the left of the 0.8 contour line conforms to the normal scaling law phenomenon. Thus, only when the minimum data size is met can the scaling law be observed (i.e., data sufficiency/surplus regime); larger models perform better. However, due to surplus data of multi-task learning and modern, optimized machine learning systems, the scaling law in LLMs can be assured in a realistic setting. Our inverse scaling phenomenon in grokking configuration can provide insight into learning a single challenging task.

\section{Grokking Mechanism of Language Models}
\label{sec:learning_process_analysis}
To better understand the phase transition between memorization and generalization in language models, we visualized the parameters of the classification layer and the L2 norm of the entire model during the learning process under the grokking configuration.
The main results are reported in Figure~\ref{fig:imdb_weights_analysis} and~\ref{fig:grokking_stages}. From the learning objective in Eq.~\ref{learning_objective}, the optimization of the model is only influenced by cross-entropy loss and weight decay.

Figure~\ref{fig:imdb_weights_analysis} showcases the dynamic evolution of the classification layer's weights. There is a sequential progression of the weights across six distinctive phases of the learning process, marked from A to F. We explain these phrases: \textbf{A}. Random Initialization: At the outset (Step 0), the weights display a high degree of variability, indicating a stochastic starting point for the learning process. \textbf{B}. Memorizing: By Step 15,000, the weights begin to show patterns of convergence (L2 norm decreases), suggesting the memorization where the model starts to fit the training data.
\textbf{C}. Memorizing 100\%: At Step 20,000, the weights begin to exhibit a regular increase, characterized by the emergence of multiple pronounced peaks. Similar results in~\cite{varma2023explaining}, the most efficient strategy for memorizing training samples is to increase its parameters. The insight of this phenomenon is to allocate distinct parameters to individual samples, reflected by the L2 norm increase.
The performance at Step 20,000 in Figure~\ref{fig:grokking_on_imdb}A indicates it has likely overfitted to the training data.
\textbf{D}. L2 Norm Decrease and Parameter Reallocation: There is a notable reduction in the amplitude of weight fluctuations, reflecting the effect of weight decay. Given that the loss approaches zero, the gradient has consequently no impact on the optimization. This phase suggests an optimization shift towards a more generalizable model~\cite{varma2023explaining}.
\textbf{E}. Grokking: Memorizing 100\%; Generalizing 85\%: As shown in Figure~\ref{fig:grokking_on_imdb}A, the model suddenly groks and achieves 85\% accuracy after overfitting. Concurrently, the parameters converge towards a fixed interval.
\textbf{F}. End Training: Under the effect of weight decay, the overall value of the parameters decreases compared to the previous stage, but they still converge at a fixed range of values.
The shift from random to fixed ranges of weights sheds light on the intricate relationship between memorization and generalization. The evolution of the L2 norm in Figure~\ref{fig:grokking_stages} also validate our explanation of the grokking phase.

\section{Case Study on Instruction Tuning}\label{sec:case_study}

To further confirm our hypothesis on a more complex task, we conduct a training dynamics analysis of the instruction tuning task under the grokking configuration. Specifically, we train a 1-layer encoder-only Transformer from scratch using the \texttt{pubmedqa\_classification} in Natural-Instruction dataset~\cite{wang-etal-2022-super}.

\begin{wrapfigure}{r}{0.6\textwidth}
    \vspace{-2em}
    \centering
    \begin{center}
        \includegraphics[width=\linewidth]{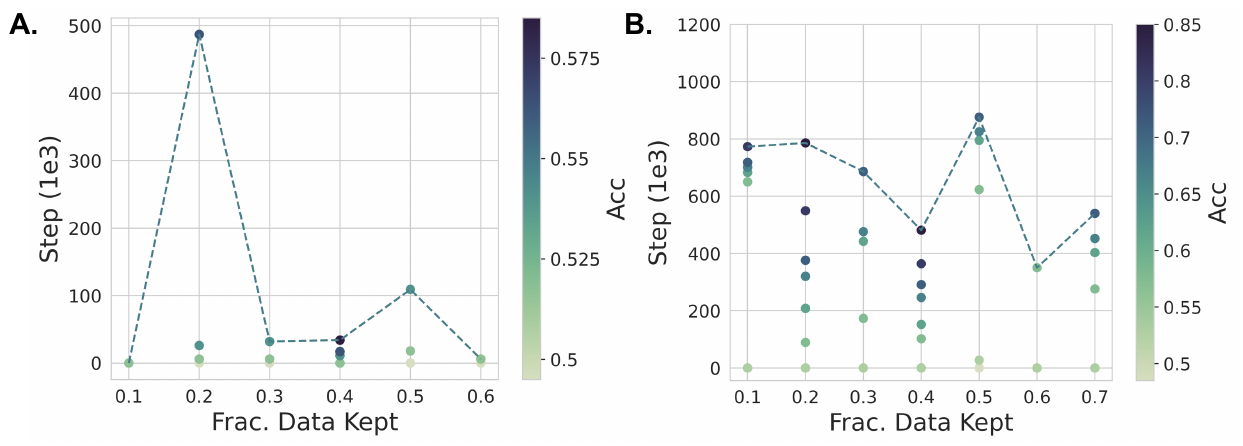}
        \caption{A case study of the instruction tuning task using a 1-layer, encoder-only Transformer. \textbf{A:} Step-wise analysis of test accuracy demonstrates clear phase transition on critical data size. \textbf{B:} Training Accuracy exhibits more fluctuations across steps, due to the complexity of the instruction tuning task.}
        \label{fig:instruction_tuning}
    \end{center}
    \vspace{-5em}
\end{wrapfigure}

Figure~\ref{fig:instruction_tuning}A demonstrates that a critical data size also exists in the complex instruction tuning task. As the dataset size increases, the model gradually transitions from slow generalization to rapid generalization, supporting our previous findings. Figure~\ref{fig:instruction_tuning}B shows the training accuracy, which generally aligns with our hypothesis. However, due to the complexity of the instruction tuning dataset, there is greater instability during training, leading to fluctuations in accuracy.


\section{Conclusion}
The main question we focus on is: How does large data drive
language models from memorization to generalization? We apply the grokking configuration to language models. 
Our analysis delves into the phase transition both from the perspectives of sample-wise and model-wise grokking during the learning process. 
We discover a critical data size that propels language models beyond mere memorization, initiating a gradual shift toward generalization. 
This leads us to establish the data efficiency hypothesis~(informal), a conceptual framework formalizing the relationship between data size and phase changes in language models.
Our data efficiency hypothesis significantly expands our understanding of how language models learn in relation to dataset size.
Further analysis indicates that the
critical dataset size increases with the model size. Language model weights undergo several regular, phased changes. A case study on instruction tuning also supports our findings.



\bibliography{example_paper.bib}
\bibliographystyle{acl_natbib}

\newpage
\appendix

\section{Related Work}\label{sec:related_wrok}
\subsection{Grokking}
Grokking~\cite{power2022grokking} presents a fascinating puzzle in the generalization problem of machine learning: neural networks start to generalize long after achieving 100\% accuracy on their training sets.
Subsequently, a series of studies have been devoted to explaining this delayed generalization phenomenon.
\citet{thilak2022slingshot} connected the grokking with the slingshot mechanism, marked by cyclic transitions between stable and unstable training. \citet{liu2022towards} developed an effective theory of representation learning, suggesting generalization originated from structured representations.
\citet{notsawo2023predicting} proposed to predict grokking using the spectral signature from the Fourier transform to detect specific oscillations in the early training phase. \citet{liu2022omnigrok} discovered that adjusting the initial scale of parameters can lead to grokking or de-grokking on the modular addition task.
By manipulating the initialization scale,
\citet{liu2022omnigrok} firstly induced the grokking on MNIST~\cite{lecun2010mnist}, IMDB~\cite{maas-EtAl:2011:ACL-HLT2011}, and QM9~\cite{ramakrishnan2014quantum} datasets with MLPs, LSTM and GCNN~(graph convolutional neural network), respectively.
\citet{nanda2023progress} demonstrated that grokking emerges not as a sudden shift, but as a gradual intensification of structured mechanisms embedded in the weights. This process is followed by the systematic elimination of memorization components. \citet{varma2023explaining} employed circuit efficiency analysis to reveal that generalization is slower to learn but more efficient. \citet{doshi2023grok} indicated that regularization methods could correct errors in the training samples.

\citet{varma2023explaining} is the closest related work, which also introduced a concept of `critical data size.'  However, the phenomenon they explain and the settings are completely different from ours. They define `critical data size' as the number of data points at which memorizing and generalizing circuits produce identical logits. Training with these data points will result in suboptimal test loss (i.e., semi-grokking). And fine-tuning grokked models with smaller data sizes will lead to poor test performance (i.e., ungrokking). 
However, we focus on the threshold that causes phase changes in language models, interpreted as the conditions under which generalization occurs. \citet{power2022grokking} and~\citet {liu2022omnigrok} also provided partial data efficiency analysis.

The above work explains grokking through the trade-off balance of memorization and generalization.
Inspired by these findings, we leverage the grokking to explore critical aspects of the phase transition in language models. 



\subsection{Data Pruning}
Recent studies on data pruning have shown that training models with key samples can maintain primary performance, comparable to those training on the full dataset. \citet{sorscher2022beyond} proposed that effective data-pruning metrics may offer a promising path for significantly improving neural scaling laws with lower resources. 
\citet{xie2023data} proposed important sampling to match the training data to target distributions.
\citet{zhou2023lima} use only 1000 samples to fine-tune a 65B parameter LLaMa~\cite{touvron2023llama} without reinforcement learning. 
They proved that using less high-quality instruction data can also achieve remarkably strong
performance.
\citet{marion2023less} presented a study on improving the quality of pretraining datasets for large language models. They discovered that using perplexity to prune training data retains main performance while reducing the dataset size by up to 70\%. 
\citet{zhu2023pad} demonstrated that smaller models, when trained on a reduced program reasoning dataset, can achieve comparable performance to LLMs.

\section{Training Details}\label{sec:training_details}
\begin{figure}[ht] 
\centering 
\includegraphics[width=0.98\linewidth]{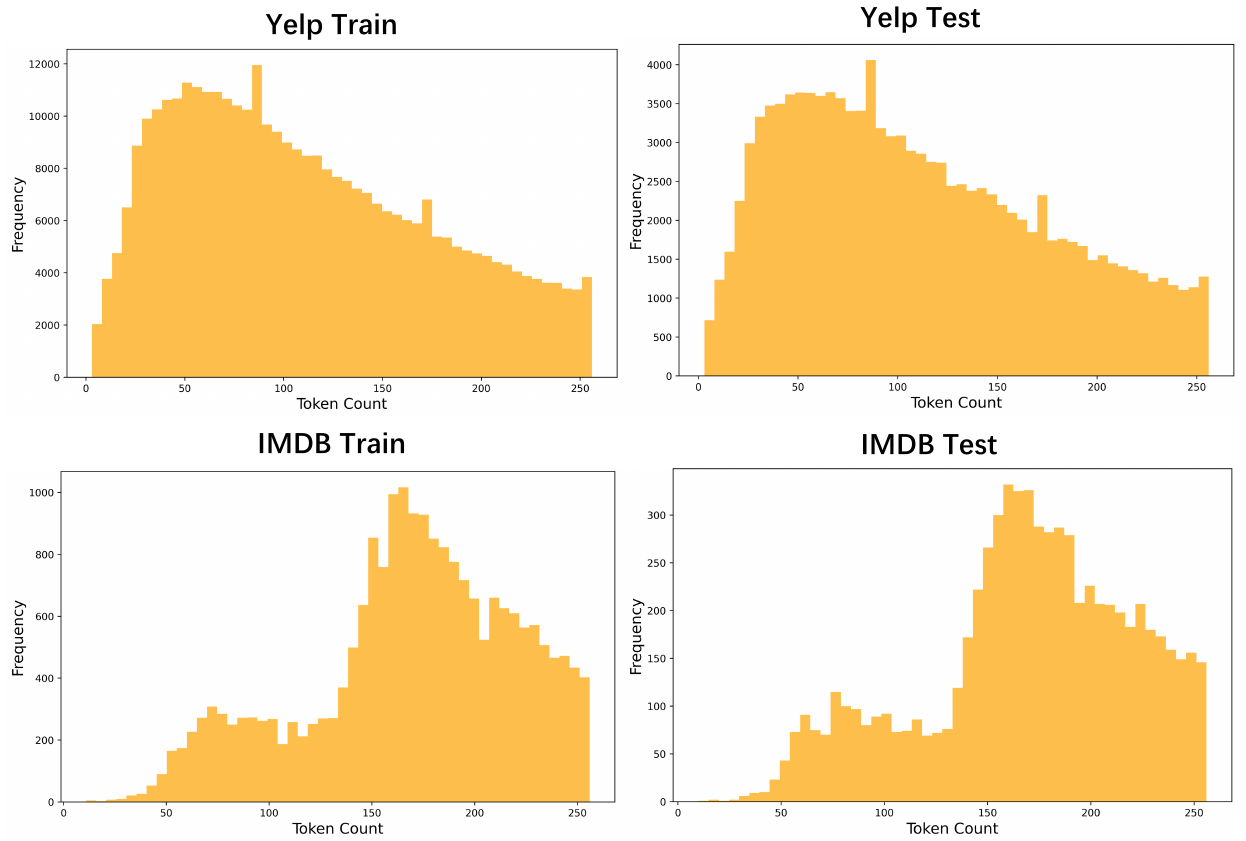} 
\caption{The statistics of the data length in the IMDB and Yelp datasets.}
\label{fig:sentiment_data_length} 
\end{figure}

\paragraph{Datasets}
We study the grokking phenomenon on modular addition~\cite{power2022grokking} and sentiment classification, with a specific case study on instruction tuning. We choose Modular Addition, IMDB and Yelp for basic hypothesis testing, because of their incrementally increasing scale. Natural-Instruction dataset~\cite{wang-etal-2022-super} is used as an more complex task for extended case study. The statistics of the datasets are detailed in Table~\ref{tab:data_statistics} and Figure~\ref{fig:sentiment_data_length}. We also provide a detailed comparison of the experimental setups between previous grokking-related papers and ours in Table~\ref{tab:grokking_datasets}. 

For modular addition, we follow~\cite{varma2023explaining} to develop a dataset in the format of
a + b \% p = n, where each character represents a token. We use the split function in Python as the tokenizer. And we set p=113, so the full data size is $p^2=12769$ and $d_{vocab}=117$. The dataset is divided into train and test sets at a 75\%:25\% ratio. For sentiment classification, we include IMDB~\cite{maas-EtAl:2011:ACL-HLT2011} and Yelp~\cite{zhangCharacterlevelConvolutionalNetworks2015} datasets. The statistics of datasets are
shown in Table~\ref{tab:data_statistics}. We import IMDB and Yelp datasets from the HuggingFace. For the instruction tuning task, we use \texttt{task846\_pubmedqa\_classification} in Natural-Instruction~\cite{wang-etal-2022-super}.
Considering the model's length constraints, we've set the maximum sample length at 256 tokens, filtering out samples exceeding this limit.
Subsequently, sample statistics and length distribution are detailed in Table~\ref{tab:data_statistics} and Figure~\ref{fig:sentiment_data_length}, respectively.
 The source datasets are from the following links:
\begin{itemize}
    \item \textbf{IMDB}:~\url{https://huggingface.co/datasets/imdb}
    \item  \textbf{Yelp}:~\url{https://huggingface.co/datasets/yelp_polarity}
    \item \textbf{Natural-Instruction}: \url{https://github.com/allenai/natural-instructions/tree/master}
\end{itemize}

\begin{wraptable}{r}{0.7\textwidth}
\vspace{-2em}
\centering
\small
\setlength{\tabcolsep}{4pt}
\caption{Statistics of the datasets used in our experiments. 
} 
\begin{tabular}{l|l|c|c}
\toprule
\textbf{Tasks}& \textbf{Datasets} & \textbf{Train} & \textbf{Test}  \\
\midrule
Modular Arithmetic & Addition & 9,576  & 3,193
\\
\midrule
\multirow{2}{*}{Sentiment Classification} & IMDB & 21,072 & 7,025
\\ 
& Yelp & 352,232 & 117,411
\\
\midrule
Instruction Tuning & Natural-Instructions & 6,500 & 650
\\
\bottomrule
\end{tabular}
\label{tab:data_statistics}
\end{wraptable}

\paragraph{Models}
For vanilla grokking, we use a 1-layer decoder-only transformer network to train from scratch. We adjust the parameters in \texttt{HuggingFace Openai-GPT config}
: the number of layers (\(n_{\text{layer}}\)) to 1, embedding size (\(n_{\text{embd}}\)) to 128, context window size (\(n_{\text{ctx}}\)) to 128, and the number of attention heads (\(n_{\text{head}}\)) to 4, vocabulary size~($vocab\_ size$) to 117, with both attention ($attn\_pdrop$) and embedding dropout ($embd\_pdrop$) rates set to 0.
We optimize the model with full batch training (i.e., using the whole training data as the training batch in each step.), using AdamW optimizer with $\beta_1=0.9,\beta_2=0.98$. The learning rate is set at 1e-3, with a weight decay of $1$. 

For grokking on IMDB, 
we use 1-layer encoder-only transformer networks to train from scratch. Specifically, we adjust the source config of \texttt{HuggingFace Bert}: the number of layers (\(num\_hidden\_layer\)) to 1, embedding size (\(hidden\_size\)) to 128, the number of attention heads ($num\_attention\_heads$) to 4, with both attention dropout~($attention\_probs\_dropout\_prob$) and hidden dropout~($hidden\_dropout\_prob$) to 0. We train the model with mini-batch $128$, using AdamW optimizer with $\beta_1=0.9,\beta_2=0.999$. The learning rate is set at 1e-3, with a weight decay of 1e-2~(default). The parameters are rescaled by the factor~$\alpha=10$. 

For grokking on Yelp and Natural-Instructions, we use the same setting as IMDB.

The source configs are from HuggingFace~\cite{wolf-etal-2020-transformers}:
\begin{itemize}
    \item \textbf{openai-gpt:}~\url{https://huggingface.co/openai-gpt/blob/main/config.json}
    \item  \textbf{bert-base-uncased:}~\url{https://huggingface.co/bert-base-uncased/blob/main/config.json}
\end{itemize}

\begin{table*}[ht]
    \centering
    \caption{Comparison of datasets and models used in previous grokking-related research and ours.}
    \resizebox{\textwidth}{!}{
    \begin{tabular}{l|c|c|c|c|c|c}
        \toprule
        \textbf{} & \textbf{Training} & \textbf{Test} & \textbf{Data type} & \textbf{Task} & Dataset & \textbf{Model} \\
        \midrule
        \cite{power2022grokking} & 7K & 2k & synthetic data & Modular Arithmetic & Addition & 1-layer Transformer \\
        \cite{thilak2022slingshot} & 7K & 2k & synthetic data & Modular Arithmetic & Addition & 1-layer Transformer \\
        \cite{liu2022towards} & 7K & 2k & synthetic data & Modular Arithmetic & Addition & 1-layer Transformer \\
        \cite{notsawo2023predicting} & 7K & 2k & synthetic data & Modular Arithmetic & Addition & 1-layer Transformer \\
        \cite{liu2022omnigrok} & 1K & 0.25k & human data & Sentiment Classification & IMDB & 2-layer LSTM \\
        \cite{varma2023explaining} & 9k & 3k & synthetic data & Modular Arithmetic & Addition & 1-layer Transformer \\
        \midrule
        \multirow{4}{*}{\textbf{Ours}}  & 9k & 3k & synthetic data & Modular Arithmetic & Addition &  1-layer Transformer \\
        & 21k & 7K & human data & Sentiment Classification & IMDB & 1/2/4/6-layer Transformer \\
        & 352k & 117k & human data & Sentiment Classification & Yelp & 1-layer Transformer \\
        & 6.2k & 0.6k & human data & Instruction Tuning & Natural-Instructions & 1-layer Transformer \\
        \bottomrule
    \end{tabular}}
    \label{tab:grokking_datasets}
\end{table*}

\begin{figure*}[ht] 
\centering 
\includegraphics[width=\linewidth]{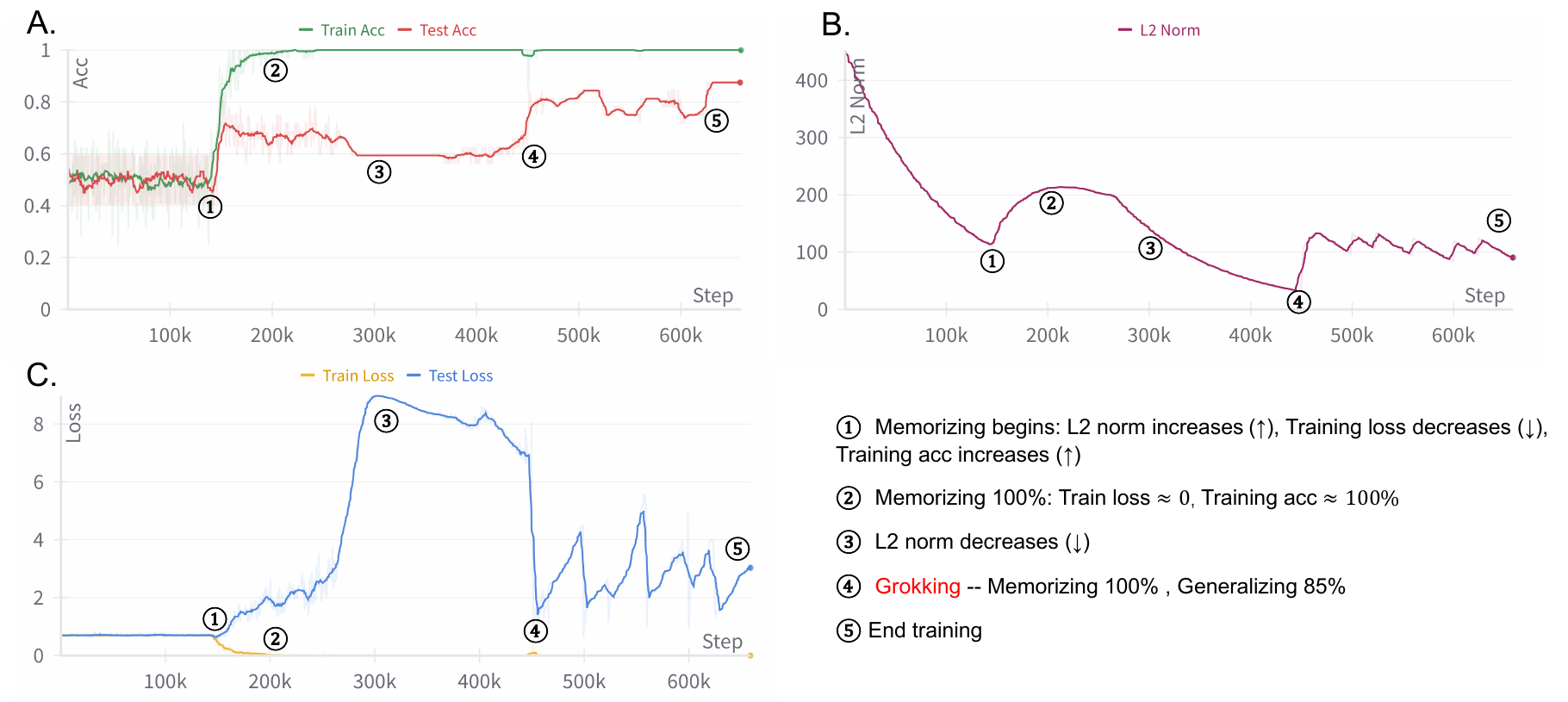} 
\caption{The evolution of the L2 norm of model parameters during the training process. Initially, the L2 norm declines driven by the loss. At \ding{192}, both the L2 norm and training accuracy increase, indicating the onset of memorization. By \ding{193}, the training loss decreases to 0 and training accuracy reaches $\approx 100\%$, signifying complete memorization. From \ding{194} to \ding{195}, the L2 norm and test loss continue to decline due to weight decay. When the L2 norm declines to \ding{195}, test accuracy rapidly shifts to $\approx80\%$, indicating grokking. From \ding{195} to \ding{196}, continued training leads to the L2 norm exhibiting constant oscillations.}
\label{fig:grokking_stages} 
\end{figure*}

\begin{figure*}[ht] 
\centering 
\includegraphics[width=\linewidth]{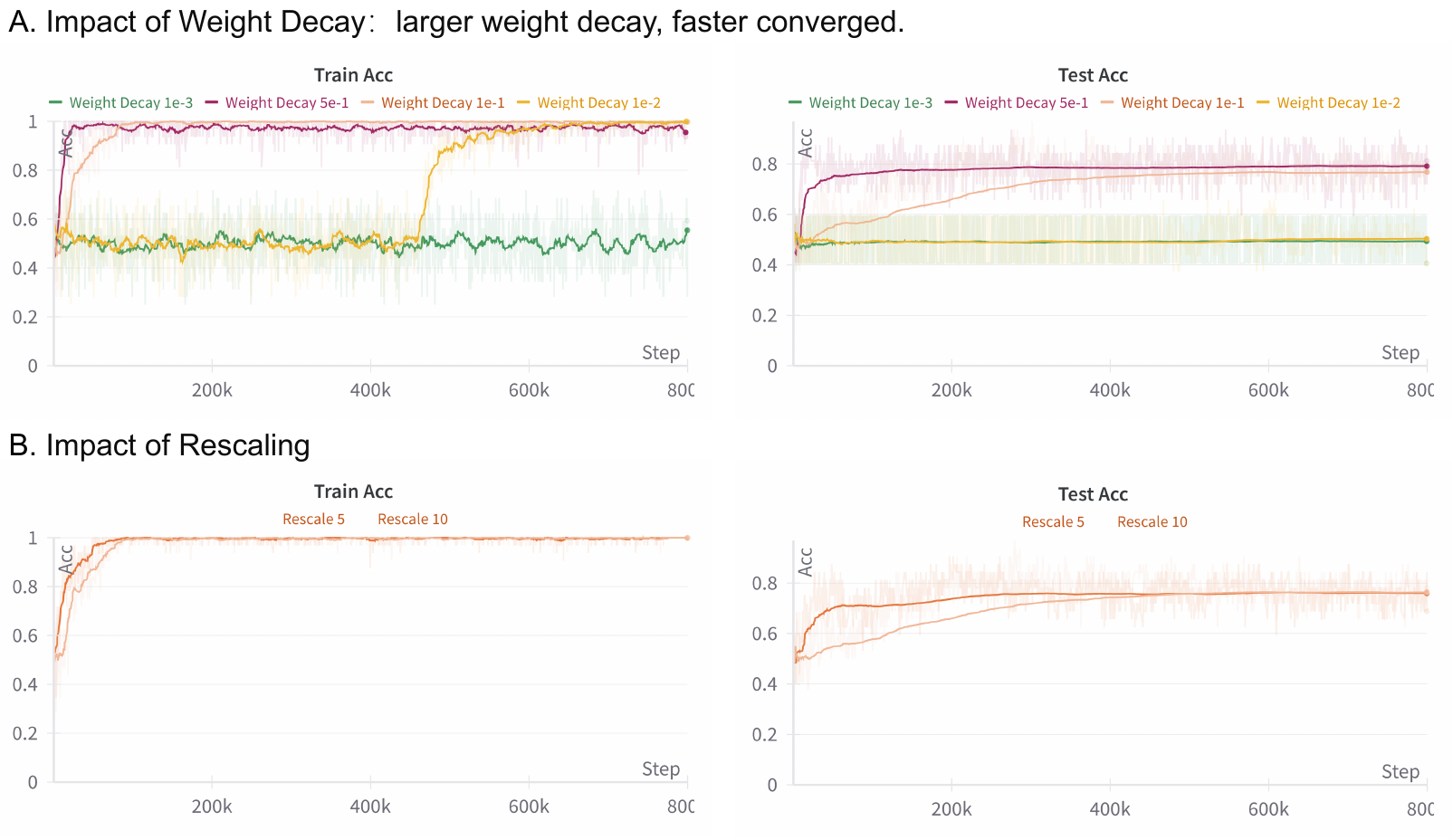} 
\caption{Impact analysis of weight decay and rescale factor: experimental results indicate that a larger weight decay accelerates convergence, while a larger rescale factor slows it down.}
\label{fig:ablation_of_wd} 
\end{figure*}

\begin{figure*}[ht] 
\centering 
\includegraphics[width=\linewidth]{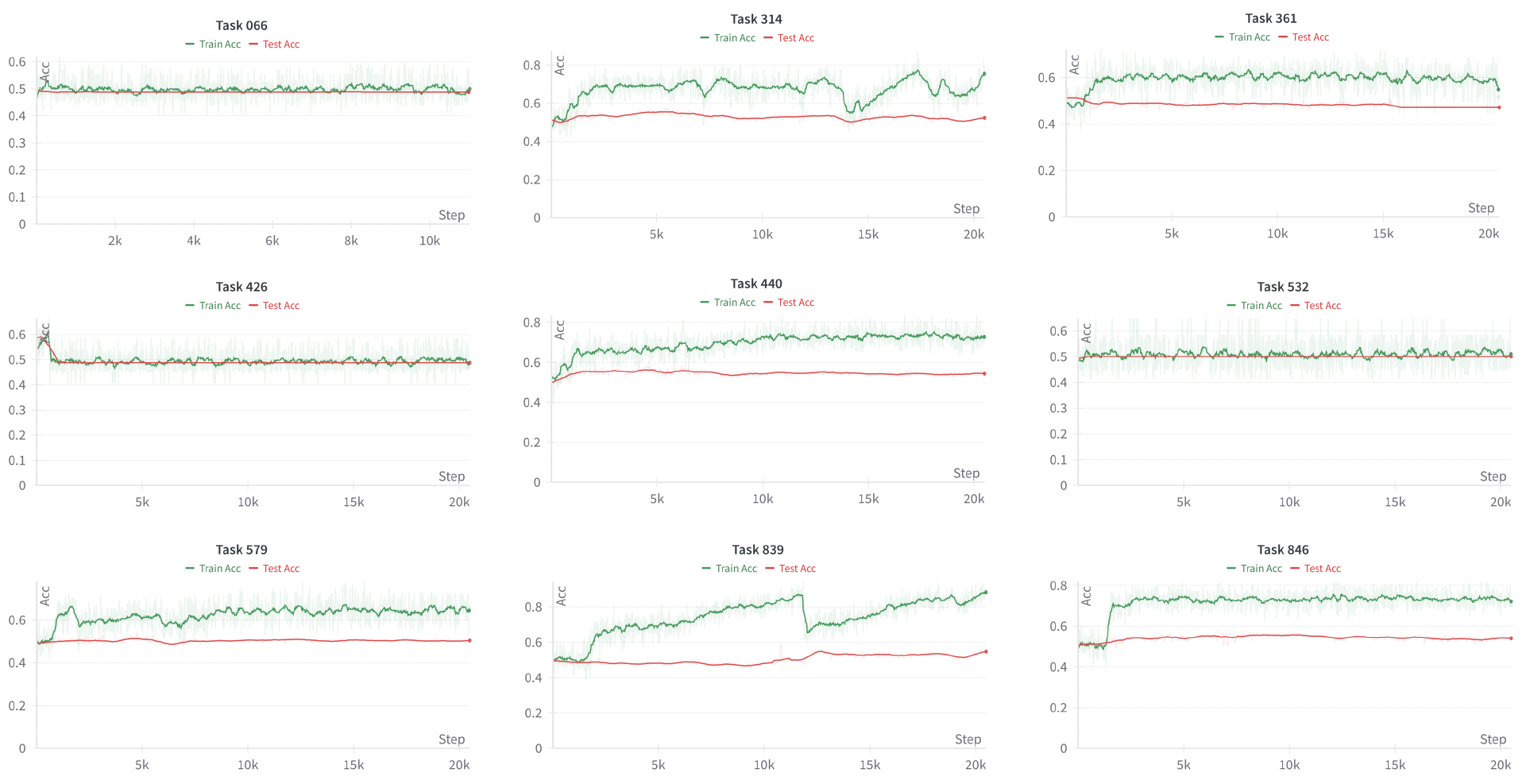} 
\caption{More training cases on instruction tuning tasks from Natural Instruction datasets~\cite{wang-etal-2022-super}.
}
\label{fig:multiple_instruction_tuning_tasks} 
\end{figure*}

\begin{figure*}[!t]
\small
  \centering
  \includegraphics[width=\linewidth]{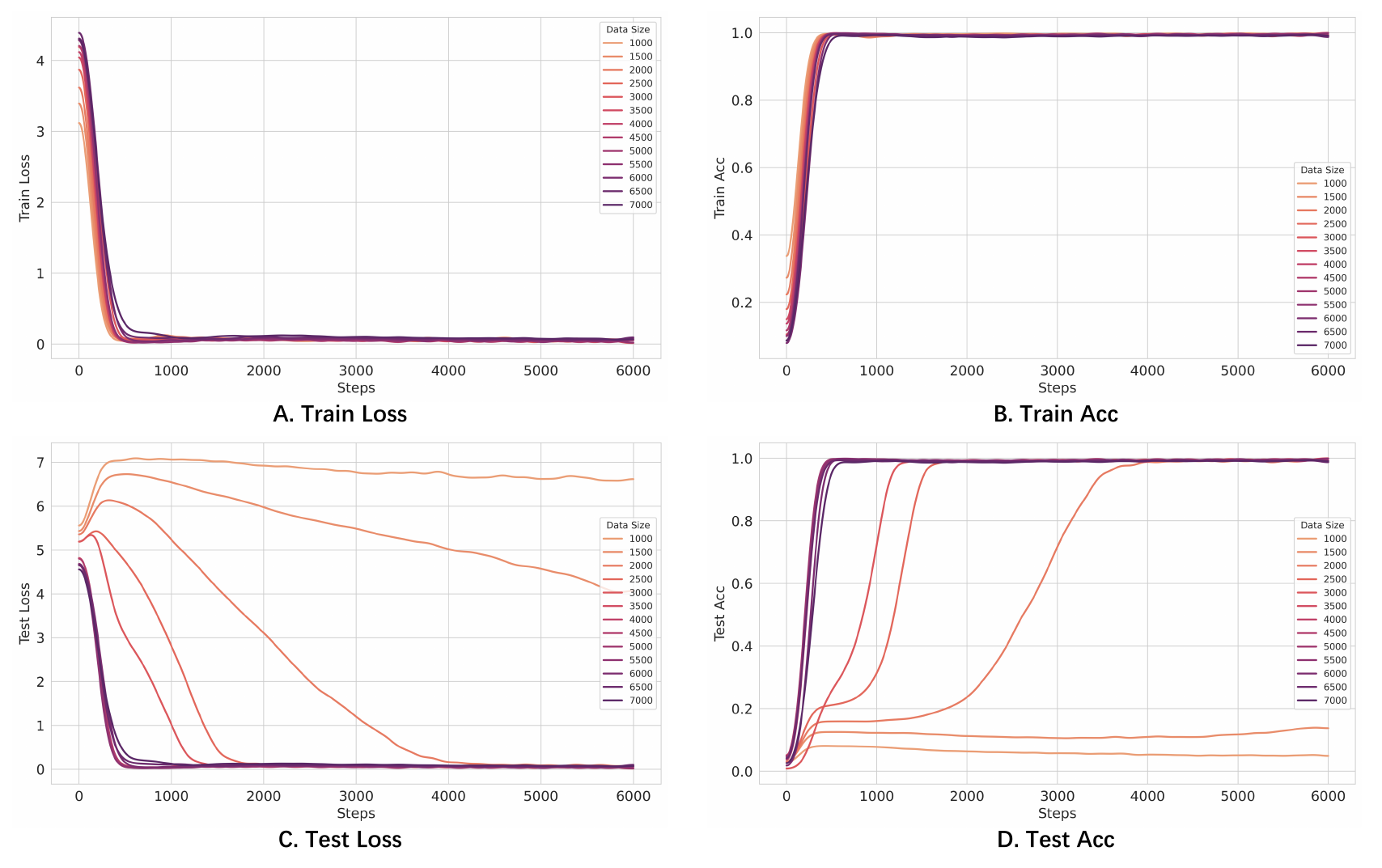}
  \caption{The detailed training process on different data sizes of modular addition under grokking setting.
  }
  \label{fig:diff_data_size_on_modular}
\end{figure*}

\begin{figure*}[ht] 
\centering 
\includegraphics[width=0.98\linewidth]{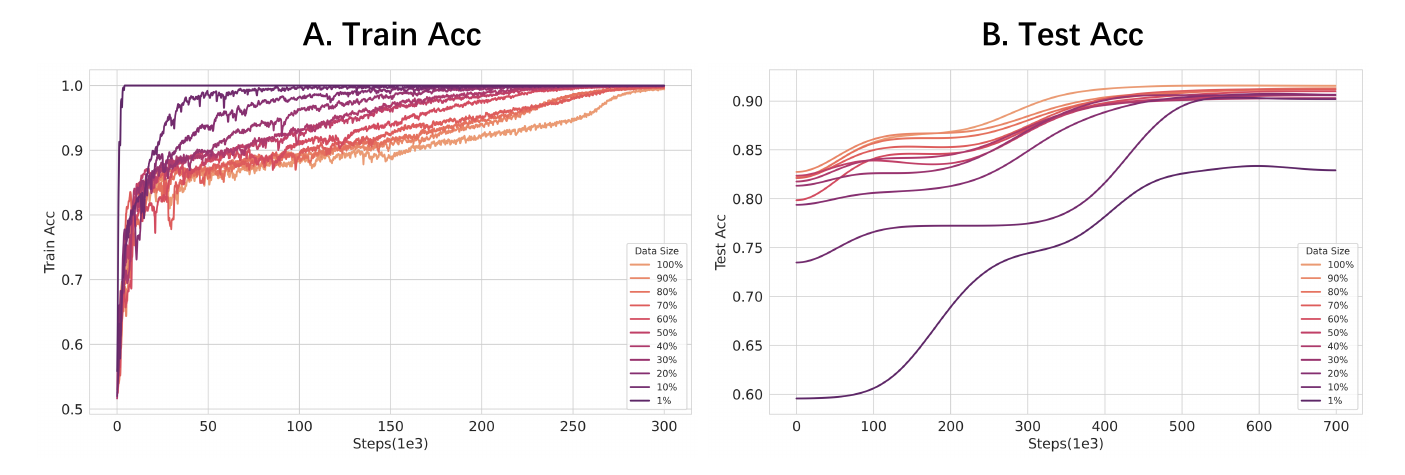} 
\caption{The training procedure on the Yelp dataset employed a 1-layer, encoder-only Transformer under the grokking framework.
}
\label{fig:acc_on_yelp} 
\end{figure*}

\begin{figure*}[ht] 
\centering 
\includegraphics[width=0.98\linewidth]{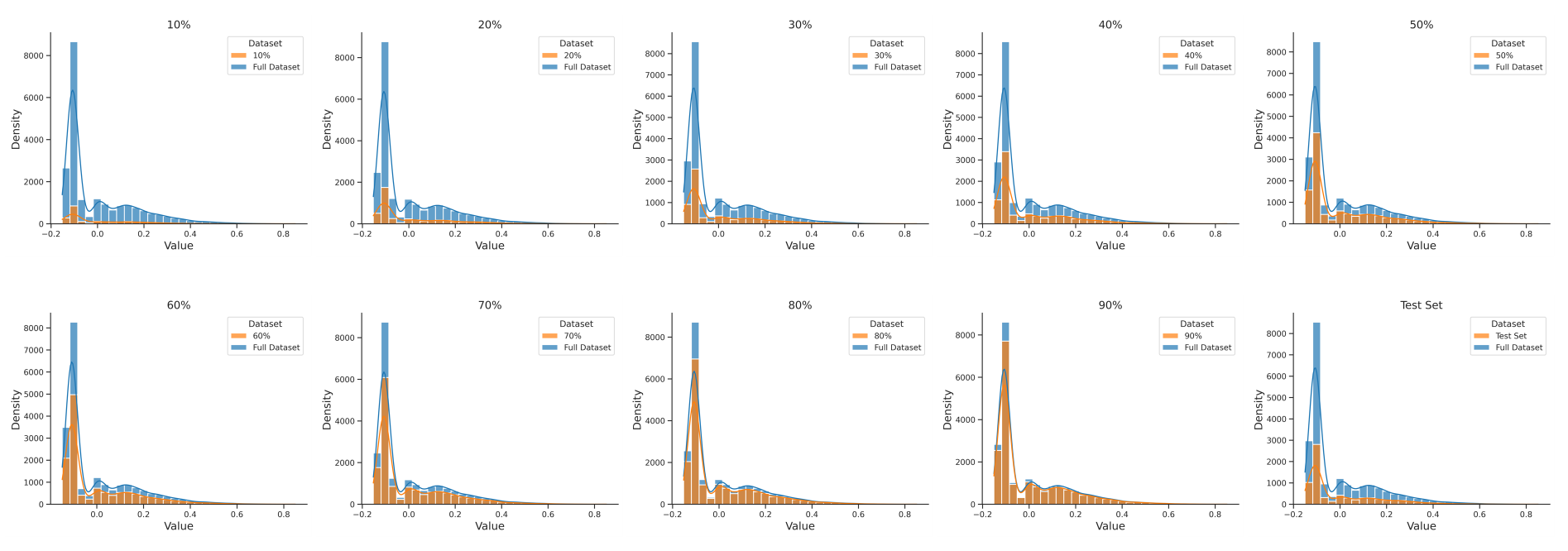} 
\caption{The 1D PCA~\cite{tipping1999mixtures} visualization of uniform data pruning on IMDB. The data pruning method uniformly removes a proportion from each principal component.}
\label{fig:data_pruning_imdb_pca} 
\end{figure*}

\section{Impact Analysis of Weight Decay and Rescaling}
 Identifying the specific values for weight decay $\lambda$ and rescale factor $\alpha$ required extensive experimentation. Inappropriate hyperparameter combinations may not induce the grokking phenomenon, as grokking requires a slower fitting speed. We supplement the ablation studies of weight decay and rescale factor in Figure~\ref{fig:ablation_of_wd}.  Ablation results illustrate larger weight decay will accelerate the convergence and a larger rescale factor will slow down the convergence.

 \section{Comparison of Experimental Settings}\label{sec:comparison_settings}
As illustrated in Table~\ref{tab:grokking_datasets}, compared to previous grokking studies, we significantly expand dataset size and scope (7-9K $\rightarrow$ 350K) within simplified language models. However, due to the fragile nature of the grokking, simultaneously enlarging both dataset and model sizes is very difficult, which requires precise control like regularization.

Our setup is currently the closest to a real-world scenario. Specifically, most previous grokking studies only focused on the Modular Arithmetic task (7-9K data) with a 1-layer Transformer (vocab\_size=115). Omnigrok~\cite{liu2022omnigrok} expanded this scope to language data by employing a 2-layer LSTM on 1K samples of IMDB. We further extend to full IMDB (21K samples) and Yelp (350K samples) datasets using a 1-layer Transformer (vocab\_size=30K, the same as BERT). However, when faced with LLMs, our current grokking configuration may struggle to slow down the learning process.

The challenge lies in that (1) adjusting weight decay may not sufficiently affect the generalization ability of LLMs to slow down convergence; (2) dataset pruning might also have minimal impact on LLMs. Thus, it will be difficult to observe such clear phase transitions. So we followed the experimental settings in \cite{power2022grokking} to vary the hidden size. As illustrated in Figure \ref{fig:model_wise_grokking}A and \ref{fig:model_wise_grokking}B, minor trend shifts are due to small changes in model sizes (i.e., from 16 to 256 hidden size). As illustrated in Figure~\ref{fig:model_wise_grokking}C, we also conducted ablation experiments on the number of layers, which similarly confirmed that larger models require larger dataset sizes to achieve generalization.

\section{Differences from Previous Work}\label{sec:Differences}
Our grokking configuration is inspired by two lines of work: (1) Grokking on the modular addition task: Grokking is associated with regularization including weight decay, dropout, and so on \cite{power2022grokking,liu2022towards,nanda2023progress}. (2) Grokking beyond the modular addition task: Grokking can be triggered on various datasets by re-scaling initialization \cite{liu2022omnigrok}.

Compared with Omnigrok \cite{liu2022omnigrok}, we have a different optimizing objective. Specifically, we dynamically adjust the initialization scale $\alpha$ and weight decay $\gamma$ to induce the grokking on classification tasks. However, Omnigrok \cite{liu2022omnigrok} used a regression task to imitate the behavior of a classification task and only rescaled initialization. Furthermore, the approach of rescaling initialization (Eq.~\ref{eq:rescale_initialization}) is derived from Omnigrok \cite{liu2022omnigrok}.

We aim to utilize grokking as a lens through which to understand the training dynamics of language models, rather than explaining grokking like previous studies (only on the modular addition task). Therefore, we conducted experiments under more realistic language datasets (Yelp, IMDB and Natural-Instructions).

\section{Training Dynamics Implementation under Data Pruning}\label{sec:Training_Dynamics}
Firstly, we utilize uniform data pruning to obtain the data fractions. Secondly, training models on data fractions, we will observe the training dynamics corresponding to data fractions. These training dynamics record the first step of arriving at the specific performance (i.e., step-wise accuracy), which contains critical features in the grokking phenomenon. For example, as depicted in Figure~\ref{fig:cds_on_modular}B, we can see a clear peak in Test Acc, indicating the beginning of generalization and the critical data size. This detailed comparison helps us to more deeply investigate the relationship between the training process and the dataset. 

\section{More Discussion}

\paragraph{Why can grokking be observed in larger language datasets ( >300k training samples, >100k test samples) beyond the toy modular addition dataset (7-9k)?}
We posit that grokking is an underlying mechanism within the model learning process. From scaling laws, we know that more samples accelerate convergence, i.e., prompt rapid generalization. Faster convergence leads to the grokking phenomenon becoming invisible in large-scale data. Regarding the IMDB and Yelp datasets in our paper, we created a “magnifying glass” to closely observe the training process by weakening regularization, increasing the scale initialization parameters, and reducing the dataset size. This approach slows down the model training process, allowing us to closely observe the phase transition within the model. Ultimately, this methodology enabled us to witness the grokking phenomenon even in the context of large datasets. However, note that these controlled conditions still only apply to simplified models; we've merely expanded the dataset size and tasks.

\paragraph{ What does grokking indicate? Does it necessitate a critical data size for its occurrence?}
Grokking can be seen as the external manifestation of a model's ability to achieve generalization on a dataset. Intuitively, there's a lot of redundancy in current datasets. Leveraging the above "magnifying glass," the grokking phenomenon suggests that the model's phase transitions may only require a critical dataset size.

\section{Limitations}\label{sec:limitations}
We utilize grokking as a lens to uncover the critical data size in language models. However, reproducing grokking on more complex tasks and larger models (e.g., 7B) is a potential challenge of our work.
Because grokking is not commonly observed in large language models with big datasets. 
As previously discussed, due to complex regularization methods and training systems, it is challenging to control LLMs to induce grokking using our current configuration. The challenges include: (1) adjusting weight decay may not sufficiently affect the generalization ability of LLMs to slow down convergence; and (2) dataset pruning might have minimal impact on these models. As a result, observing clear phase transitions is difficult. Additionally, the use of toy settings is also a typical issue in grokking-related papers. However, this paper is still a good start in applying the insights of grokking to language models. We will further address these issues in future work.

\begin{wrapfigure}{r}{0.7\textwidth}
    \vspace{-2em}
    \centering
    \begin{center}
\includegraphics[width=0.98\linewidth]{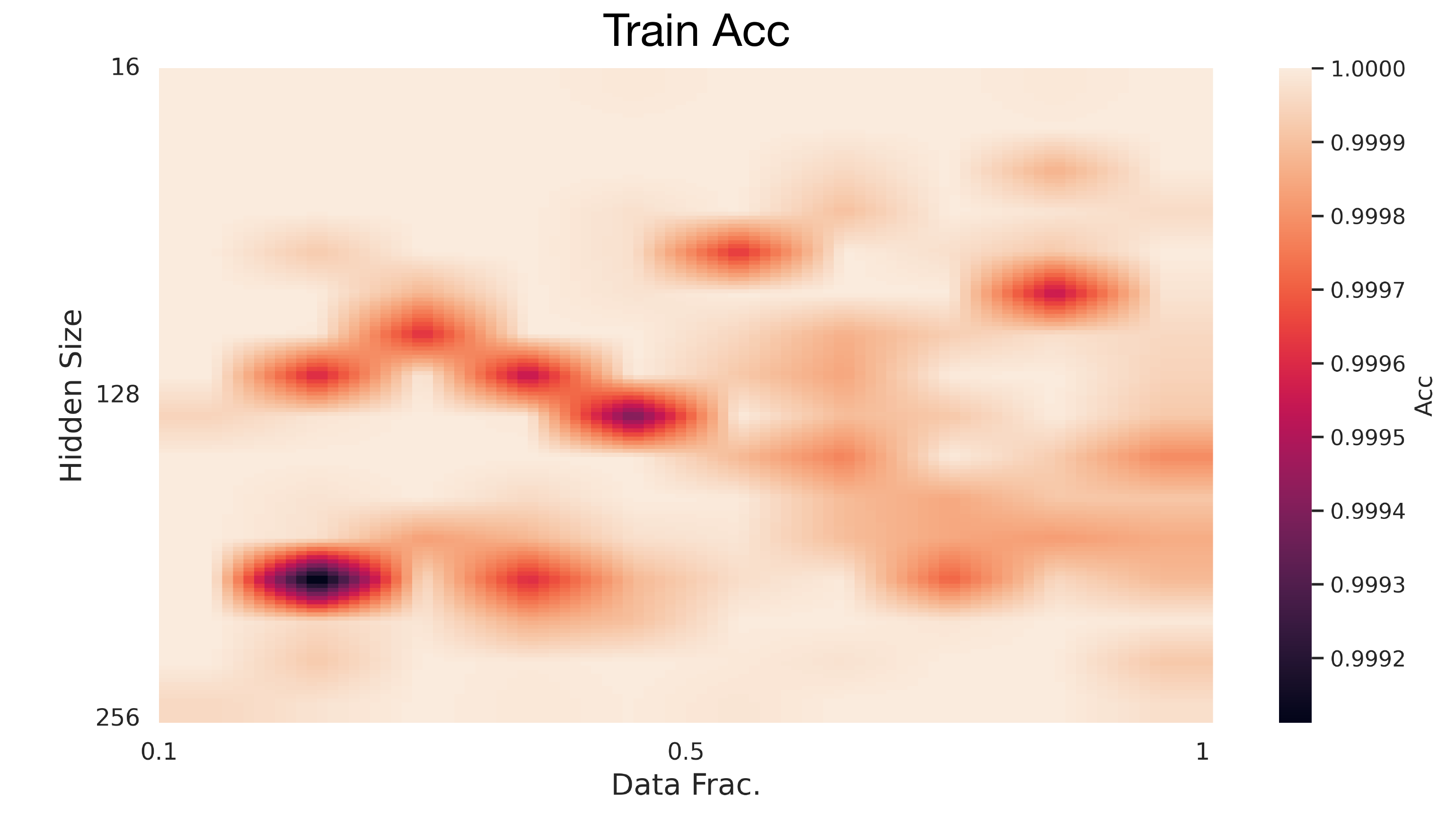} 
\caption{Training accuracy as a function of hidden size and data fraction. 
}
\label{fig:model_wise_grokking_train_acc} 
\end{center}
\end{wrapfigure}

\section{Future Work}
In our existing results, we have explored vast sizes of datasets. However, the challenge with LLMs is their stronger generalization capability and rapid fitting speed. Due to the unstable grokking across different settings, inappropriate hyperparameter combinations may not induce the grokking phenomenon, as grokking requires a slower fitting speed. 
Grokking phenomenon requires regularized representational learning and weakened regularization conditions, a process that necessitates careful control. 
We provide detailed comparisons of experimental settings in grokking-related papers in Table~\ref{tab:grokking_datasets}.

We will include this in our future directions. 
Below, we'll discuss in detail how we plan to tackle this challenge.
Specifically, we will explore the following two directions of critical data size in LLMs:
\begin{itemize}
    \item How to control the convergence speed under LLMs conditions, then carefully observe the phase changes during the training process ? Our methods include weakening regularization and dataset pruning.
    \item We will explore the representation dynamics on more complex tasks (e.g., unsupervised pre-training, instruction tuning ). We aim to seek the phase change effects between memorization and generalization in LLMs.
\end{itemize}

\section{Broader Impacts}
\label{sec:broad}
Language models carry potential risks, such as generating offensive language, reinforcing social biases, and leaking private information. However, our research can deepen the understanding of language models and help prevent potential threats. Therefore, we believe that the benefits of open-sourcing our models and research significantly outweigh the potential downsides.

\end{document}